# The Hidden Web, XML and the Semantic Web: Scientific Data Management Perspectives

Fabian M. Suchanek *
INRIA Saclay
Paris
France
fabian@suchanek.name

Aparna S. Varde
Dept. of Computer Science,
Montclair State University
Montclair, NJ
USA
vardea@montclair.edu

Richi Nayak
Computer Science
Discipline, Queensland
University of Technology
Brisbane, Australia
r.nayak@qut.edu.au

Pierre Senellart *
Institut Télécom; Télécom
Paristech; CNRS LTCI
Paris
France
pierre@senellart.com

## ABSTRACT
The World Wide Web no longer consists just of HTML pages. Our work sheds light on a number of trends on the Internet that go beyond simple Web pages. The hidden Web provides a wealth of data in semi-structured form, accessible through Web forms and Web services. These services, as well as numerous other applications on the Web, commonly use XML, the eXtensible Markup Language. XML has become the *lingua franca* of the Internet that allows customized markups to be defined for specific domains. On top of XML, the Semantic Web grows as a common structured data source. In this work, we first explain each of these developments in detail. Using real-world examples from scientific domains of great interest today, we then demonstrate how these new developments can assist the managing, harvesting, and organization of data on the Web. On the way, we also illustrate the current research avenues in these domains. We believe that this effort would help bridge multiple database tracks, thereby attracting researchers with a view to extend database technology.

## Categories and Subject Descriptors
H.2.8 [Database Management]: Database Applications – Scientific databases, H.3.3. [Information Search and Retrieval] – Search process

## 1. INTRODUCTION
### 1.1 Background and Motivation
The Web has revolutionized the way we seek information. It covers domains as diverse as education, entertainment, business, and even health. It has evolved into a publishing medium, a global information repository and a global electronic market. From a user's perspective, the Internet is mainly a source of Web pages. Yet, there is much more to the Web than HTML pages. Much of its content lies in databases behind the Web. This content, called the hidden Web, is accessible only through Web forms or through specific Web services. It is estimated that the hidden Web contains orders of magnitude more data than the visible Web. Yet, despite its providing structured data, it is not obvious how to access this data by a machine. In order to make use of this wealth of information, Web forms have to be found and filled in, and the requested answers have to be analyzed and understood. Through the medium of this tutorial, we show the newest research trends in the area of the hidden Web and tell the story of various early successes. We also focus on the Web developments in two other exciting areas, namely XML and the Semantic Web.

The increasing use of XML is an interesting trend on the Web. XML, the Extensible Markup Language, is a W3C standard for the exchange of semi-structured data. Through XML, it has become possible to transfer data between systems as diverse as databases, Web services, semantic knowledge bases or end user applications in one common file format. Yet, XML by itself defines just the abstract syntax, the hull of the information. It does not standardize the semantics or the structure of the data. This is the role of domain-specific markup languages (DSMLs). In this tutorial, we explain the design principles of DSMLs and how they blend into XML. We first give an overview of the most established DSMLs, thereby diving into the realm of real-world XML as it is used on the Web. We then also show how various mining techniques can be applied to XML data for better knowledge discovery.

The latest development on the Web is the Semantic Web – the newly growing structure of computer-processable meaningful information. With its W3C standards RDF and OWL, the Semantic Web aims to unify the way semantic information is stored and exchanged, making it possible for one system to not just read, but also "understand" the data from another source. Our tutorial provides basic literacy with these standards. We also show where the Semantic Web has already taken off – with example knowledge bases from the science domain, the music domain or indeed general common sense knowledge. Still, the Semantic Web is in its infancy. Numerous challenging research questions, e.g., in areas of information reconciliation and knowledge representation, still long to be answered.

* The work was partially funded by the ERC grant Webdam.

## 1.2 Goals

The overall goal of this tutorial is to show how all these diverse trends can be used for data management. As an example, we choose the scientific domain. Interdisciplinary research has attracted tremendous interest in recent years with the emerging need for computer scientists to work across domains. This has led to significant development of techniques and standards that use computational principles to solve challenging real-world problems in specific domains, with new theories being proposed as needed. In the last decade, a large part of this research has occurred in the field of scientific domains. Management of scientific data has accordingly become an exciting interdisciplinary field of study. Moreover, in this age of globalization, it is imperative to share the information with communities of scientists and related professionals from all over the world through the Web. We therefore shed light on new methods of scientific data management on the Web. On the way, we illustrate the various research questions in these domains that are still unanswered.

We aim to use various examples from scientific data throughout including markup languages specifically developed for scientific domains. For instance, we show how data about Cédric Villani (recipient of the Fields Medal 2010) and his works can be stored and retrieved using each of these technologies. Data on his research, not so obvious from the surface Web, can be found using intentional and extensional approaches in the hidden Web. Meaningful data extraction about him, including personal information that requires some human reasoning, can be carried out using Semantic Web technologies. XML and its domain-specific markup languages such as MathML, can store additional information, such as mathematical data, in an expressive Web-based format.

These technologies are by no means finalized. This tutorial thus endeavors to highlight the open issues in these domains and point out the hot spots of today's research. It would thus encourage future work across these areas, opening new avenues for research and development, especially propelling further research in areas that blend these technologies.

## 1.3 Outline

This tutorial is organized in three parts, each one focusing on one aspect of scientific data management. The first part delves into the hidden Web (also known as the deep Web or invisible Web), explaining how it can be used for storing and retrieving scientific information. The second part focuses on XML data management along with domain-specific markup languages (DSMLs), particularly with respect to modeling and mining scientific data. The third part presents the details of the Semantic Web with emphasis on its usefulness for managing scientific data.

## 2. SCIENTIFIC DATA MANAGEMENT

### 2.1 The Hidden Web

It is impossible to reach the whole content of the World Wide Web by just following hyperlinks. Rather, some Web pages create their content on demand if the user fills out and submits a Web form. These forms are typically interfaces to databases stored on Web servers, which are usually not directly accessible. A reason behind this may be that the Web site owner wants to keep control on how the information is accessed. Some hidden Web pages are "Yellow pages" services and other kinds of directories; others are library catalogs, weather forecast services, geo-localization services, commercial product catalogs or administrative sources such as the database of the United States census bureau.

As an example in the scientific data management domain, let us consider the following task: "Retrieve all articles published by Cédric Villani." This information is most likely available from the surface Web, but a classical Web crawler would have to explore a large number of documents, and then use some heuristics or classification techniques to determine the relevant ones. This is a time-consuming and imprecise process. On the other hand, using services of the hidden Web, this becomes a very simple task: simply use the advanced querying capabilities of a publication database, such as Google Scholar.

The question is how users can benefit from this vast source of information. How can they discover it, index it, and be able to use it automatically for a given task? We review some ways to tackle these challenges. We focus here especially on unsupervised and semi-supervised approaches for knowledge discovery over the hidden Web such as [11], more adapted to the scale and diversity of the Web than supervised methods. Most current approaches can be roughly classified [9] into *extensional* or *surfacing* strategies (retrieving information from the hidden Web and storing it locally to process it) and *intentional* or *integration* strategies (analyzing services to understand their structure, store this description, and use it to forward users' queries to the services). By covering these two avenues, we give an overview of the most important current approaches of data management on the hidden Web.

### 2.2 XML and DSMLs

With the continuous growth in XML-based data sources, especially in scientific applications, the ability to manage collections of XML documents becomes increasingly important [7]. In such applications, it is useful to capture the reasoning process of domain experts, especially in modeling and mining the data. Modeling of XML documents, which are semi-structured, requires finer details than unstructured (text) documents and fully structured documents. XML allows the representation of semi-structured and hierarchal data containing not only the values of individual items but also the relationships between data items. Due to the inherent flexibility of XML, in both structure and semantics, mining of XML documents significantly differs from structured data mining and text mining [6]. Consequently, management of XML data is faced with new challenges as well as benefits [10,14].

A considerable advancement in the general area of XML is the advent of domain-specific markup languages (DSMLs). These languages are designed such that they follow the syntax of XML and encapsulate the semantics of the concerned domain in order to cater to the needs of targeted user bodies. Accordingly, they serve as communication standards in the respective domains. The storage of data in this format has several advantages such as more meaningful information retrieval and knowledge discovery. This is particularly useful in scientific applications where it is crucial to incorporate the domain experts' perspective in various real-world situations.

We discuss the issues in managing XML-based scientific data. We address XML modeling and mining, focusing on both structure as well as content of XML documents. We give details such as vector, tensor, tree and graph structure representation for XML data, and their usefulness in scientific data management. We cover DSMLs with emphasis on scientific domains, giving illustrative examples from MML (Medical Markup Language) [5] and MathML (Mathematical Markup Language) [3]. We explain the steps in DSML development [15], the desired properties of DSMLs, and the application of XML constraints to preserve semantic restrictions [4].

Furthermore, we describe how the XML model used for data storage facilitates information retrieval with languages in the XML family such as XQuery [2]. We also explain the advantages of using classical data mining techniques such as association rules and clustering over XML documents in conjunction with DSMLs to enhance knowledge discovery, especially from scientific data. Suitable examples from domains such as medicine are provided here [8,15]. Finally, open research questions are addressed pertaining to structure and content mining and a greater synergy between XML and DSMLs.

## 2.3 Semantic Web Technologies

The Semantic Web project envisions that people will publish semantic information in a computer-processable formalism that allows the information to be globally interlinked. For this purpose, the World Wide Web Consortium (W3C) has developed the knowledge representation formalisms RDF [18] and OWL [17]. These formalisms are based on XML, but go beyond it by specifying semantic relationships between entities and even logical constraints on them. A collection of world knowledge in these formalisms is commonly called an *ontology* [12]. In this tutorial, first give an introduction to semantic knowledge representations and ontologies in general. We also explain the knowledge representation formalisms RDFS and OWL, their syntax and semantics. We explain the vision and the applications of the Semantic Web project.

We then show how the Semantic Web can be used for scientific data management. We introduce different existing semantic resources for scientific data, from publication databases such as DBLP, which have been published in RDF, to general purpose ontologies such as YAGO [13] or DBpedia [1], each containing thousands of scientists and scientific concepts and theories. Several large-scale ontologies are available online and are interlinked in the spirit of the Semantic Web.

We also explain how this information was gathered from different sources and how it can be queried using the SPARQL query language [19].

## 3. CHALLENGES

We highlight some of the challenging issues leading to current research questions in each of these Web developments addressed in this work.

In the area of the hidden Web, although there are techniques to derive the type of input parameters and output records of a Web form, it is still challenging to get the precise semantics of a service. If a service, given a person, returns a year, how do we know whether this is the birth date, death date, or graduation date? Moreover, dealing with complex forms (such as those used to access specialized scientific databases), especially when there are dependencies between form fields, required and optional fields, etc., is a fully open problem by itself. Finally, a knowledge discovery system for the hidden Web that would not cover specific Web sites or a specific domain of interest but would work at the scale of the whole Web is still to be constructed.

In the area of XML and DSMLs, some of the current issues and challenges in managing scientific data are:

1. Effectively modeling both structure and content features for XML documents to adequately represent scientific data and investigating how DSMLs can be useful here

2. Combining structure and content features in different types of data models which do not affect the *scalability* of the mining process

3. Integrating background knowledge of scientific processes in XML mining *algorithms* and harnessing DSMLs here

4. Developing new standards as needed especially to foster knowledge discovery by synergizing XML and DSMLs

The Semantic Web is still relatively young. Numerous challenges still wait to be solved:

1. One of the main open research questions is the reconciliation of different semantic conceptualizations in different ontologies. Techniques such as record linking and matching carry over from the database world, but present themselves in the new light of the Semantic Web, with more information and constraints available to support the mapping process.

2. Another challenge is the growing of the Semantic Web – be it through community work, by converting existing databases into RDF, or by Information Extraction.

3. Reasoning on Web scale is likewise still an open issue. How can we apply automated reasoners on huge, potentially noisy data sets?

4. The Semantic Web can also be mined, for example to find sub- and super-class relationships or to find schema information.

With the presentation of these challenges, we aim at encouraging even more research in the areas of the hidden Web, XML, and the Semantic Web, to further enhance scientific data management.

## 4. CONCLUSIONS

This tutorial would show that developments in the areas of the hidden Web, Semantic Web, XML and DSMLs have influenced the current state of the Web by improving the way different computer applications and services communicate, and relevant information is identified, especially in scientific domains.

Most of the data on the Web is so unstructured that only humans can understand it. However, the volume of the data is so huge that only machines can process it. This paradox highlights the importance of techniques "in between" the extremes: semi-structured languages such as XML, the Semantic Web and the hidden Web. This tutorial sees itself as an invitation to researchers to make use of these technologies on one hand – and to contribute to these developments on the other hand. We believe that there is exciting research potential in bridging the different new developments on the Web.

## 5. AUTHOR BIOGRAPHIES

**Fabian Suchanek** is a postdoctoral researcher at INRIA Saclay in Paris. He spent one year as a visiting researcher at Microsoft Research Silicon Valley. Fabian obtained his doctoral degree at the Max Planck Institute for Informatics/Germany in 2008. In his dissertation, Fabian developed methods for the automatic construction and maintenance of a large knowledge base, YAGO. For his thesis, he received the ACM SIGMOD Dissertation Award Honorable Mention. The original YAGO paper at the WWW Conference in 2007 has received more than 350 citations, and YAGO is used in many major knowledge-base projects around the world (including DBpedia).

**Aparna Varde** is a Tenure Track Assistant Professor in Computer Science at Montclair State University, NJ. She obtained her Ph.D. in the area of scientific data mining from Worcester Polytechnic Institute, MA. For her work, she was given an Associate membership of Sigma Xi, the Scientific Research Society for excellence in multidisciplinary research. Her research interests span data mining, databases and artificial intelligence with particular emphasis on scientific domains. This has lead to over 35 publications in these areas. She has served on the PC of various conferences, e.g., ER 2010, EDBT 2009, ICDM 2008 and been a reviewer for journals including TDKE, DKE and the VLDB journal. She is advising a Ph.D. student and has been the research advisor for 4 M.S. students and 5 B.S. students. She is supported through various grants by organizations such as the National Science Foundation, USA.

**Richi Nayak** is a senior lecturer in Computer Science discipline at Queensland University of Technology, Brisbane, Australia. Her research interests are data mining and knowledge discovery. In recent years she has focused her research on Web intelligence, Web service discovery and XML data management. She has been successful in applying theories of data mining in a variety of practical scientific settings. Her publications include 1 edited workshop proceeding, 1 edited book, 8 book chapters, 15 refereed journal articles and 60 refereed conference articles. She is founder and chief editor of the International Journal of Knowledge and Web Intelligence.

**Pierre Senellart** is an Associate Professor in the Computer Science and Networking department at Télécom ParisTech, the French leading engineering school specialized in information technology. He is an alumnus of the École normale supérieure and obtained his Ph.D. (2007) in Computer Science from Université Paris-Sud, studying under the supervision of Serge Abiteboul. He has published research in top-tier computer science conference and journals and has served on the PC and as an officer of various conferences; he is also the Information Director of the *Journal of the ACM*. His research interests focus around theoretical aspects of database management systems and the World Wide Web, and more specifically on the intentional indexing of the hidden Web, probabilistic XML databases, and graph mining.